\documentclass[conference]{IEEEtran}
\IEEEoverridecommandlockouts
\usepackage{cite}
\usepackage{amsmath,amssymb,amsfonts}
\usepackage{algorithmic}
\usepackage{graphicx}
\usepackage{textcomp}
\usepackage{xcolor}
\usepackage[small]{caption}
\usepackage{balance}
\usepackage{times}
\usepackage{soul}
\usepackage{url}
\usepackage{amsthm}
\usepackage{booktabs}
\usepackage{algorithm}
\usepackage{algorithmic}
\usepackage{makecell}
\usepackage{multirow}
\usepackage{bm}
\usepackage[stable]{footmisc}
\theoremstyle{definition}

\usepackage{mathrsfs}
 \usepackage{enumitem}
 \usepackage{romannum}
 \usepackage{lipsum}
\newcommand{\R}[1]{{\color{black}{#1}}}
\newcommand{\ie}{\emph{i.e., }}

\usepackage{newunicodechar}

\def\method{DyHSL}

\def\BibTeX{{\rm B\kern-.05em{\sc i\kern-.025em b}\kern-.08em
    T\kern-.1667em\lower.7ex\hbox{E}\kern-.125emX}}

\title{Dynamic Hypergraph Structure Learning for Traffic Flow Forecasting}

\author{
\IEEEauthorblockN{Yusheng Zhao$^{1*}$, Xiao Luo$^{2*}$, Wei Ju$^{1,\dagger}$, Chong Chen$^{3}$, Xian-Sheng Hua$^{3}$, Ming Zhang$^{1,\dagger}$}
\IEEEauthorblockA{$^1$\textit{School of Computer Science, Peking University}, \\ $^2$\textit{Department of Computer Science, University of California Los Angeles}, $^3$\textit{Terminus Group} \\
$^1$yusheng.zhao@stu.pku.edu.cn, $^2$xiaoluo@cs.ucla.edu, $^1$\{juwei, mzhang\_cs\}@pku.edu.cn,
\\
$^3$cheung1990@126.com, $^3$huaxiansheng@gmail.com}
\thanks{$^*$ Equal contribution with order determined by flipping a coin.}
\thanks{$^\dagger$Corresponding authors.}
}

\begin{document}
\maketitle

\begin{abstract}
This paper studies the problem of traffic flow forecasting, which aims to predict future traffic conditions on the basis of road networks and traffic conditions in the past. The problem is typically solved by modeling complex spatio-temporal correlations in traffic data using spatio-temporal graph neural networks (GNNs). However, the performance of these methods is still far from satisfactory since GNNs usually have limited representation capacity when it comes to complex traffic networks. Graphs, by nature, fall short in capturing non-pairwise relations. Even worse, existing methods follow the paradigm of message passing that aggregates neighborhood information linearly, which fails to capture complicated spatio-temporal high-order interactions. To tackle these issues, in this paper, we propose a novel model named \underline{Dy}namic \underline{H}ypergraph \underline{S}tructure \underline{L}earning (DyHSL) for traffic flow prediction. To learn non-pairwise relationships, our DyHSL extracts hypergraph structural information to model dynamics in the traffic networks, and updates each node representation by aggregating messages from its associated hyperedges. Additionally, to capture high-order spatio-temporal relations in the road network, we introduce an interactive graph convolution block, which further models the neighborhood interaction for each node. Finally, we integrate these two views into a holistic multi-scale correlation extraction module, which conducts temporal pooling with different scales to model different temporal patterns. Extensive experiments on four popular traffic benchmark datasets demonstrate the effectiveness of our proposed DyHSL compared with a broad range of competing baselines.

\end{abstract}



\begin{figure}[t]
\centering
\includegraphics[width=0.48\textwidth,keepaspectratio=true]{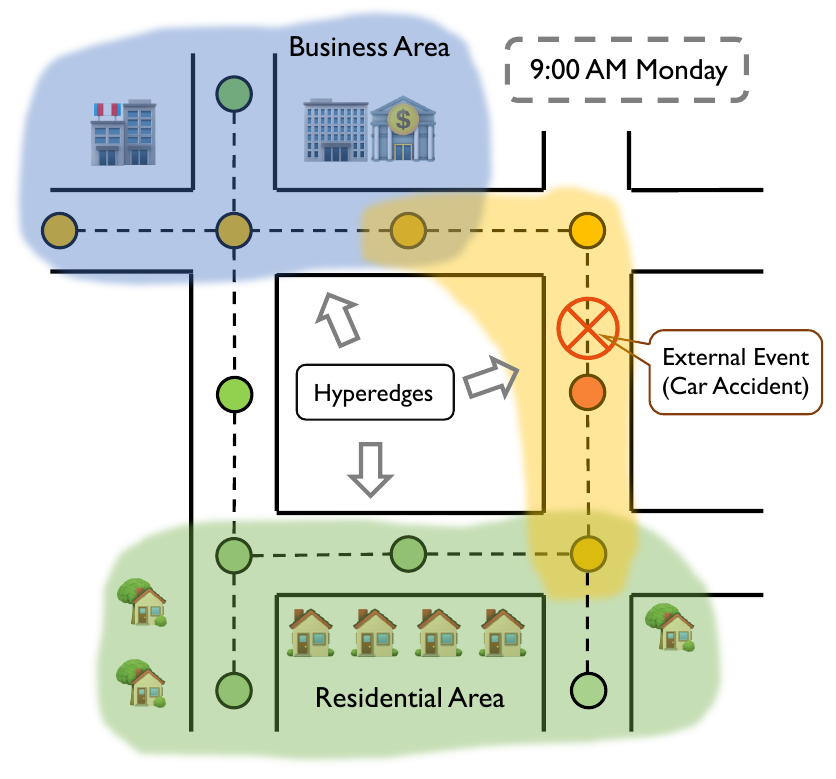}
\caption{An illustration of the dynamic hypergraph structure in the traffic network. The business area and the residential area both could imply static hyperedges whereas external events could bring in dynamic hyperedges. 
}
\label{fig:motivation}
\end{figure}

\section{Introduction}

Spatio-temporal forecasting~\cite{wu2021autocts} has been a basic topic with a range of applications including traffic flow forecasting~\cite{ji2022stden}, physical law analysis~\cite{wangphysics,huang2021coupled} and disease spreading understanding~\cite{zheng2021hierst}. Among various related practical problems, traffic flow forecasting aims to predict future traffic conditions on the basis of road networks and traffic conditions in the past~\cite{li2022dynamic}. This problem plays an important role in urban systems, which can significantly benefit congestion management.

In literature, a fruitful line of traffic flow forecasting methods has been developed, which can be roughly divided into physics-based methods and learning-based methods. Typically, physics-based methods leverage coupled differential equations to characterize traffic systems~\cite{ni2015traffic,di2021survey}. They usually achieve superior performance in simulated data with theoretical guarantees. However, they often rely on a strong model assumption, which is difficult to meet in complicated situations of the real world~\cite{mo2021physics}. In contrast, learning-based methods attempt to utilize historical observations to optimize a machine learning model, which has been popular among various solutions. Early efforts attempt to incorporate traditional models such as autoregressive integrated moving average~\cite{williams2003modeling} (ARIMA) and support vector machine~\cite{sun2014traffic} (SVM) into this problem. Recently, deep learning-based methods have achieved better performance benefiting from the representation capacity of deep neural networks.
On the one hand, these methods usually utilize graph neural networks~\cite{kipf2016semi} (GNNs) to extract structured spatial relationships from road networks. On the other hand, they utilize recurrent neural networks~\cite{2017DCRNN} (RNNs) or temporal convolution networks~\cite{GWNet} (TCNs) to extract temporal relationships. By integrating different networks into spatio-temporal GNNs~\cite{2020GMAN, 2020MTGNN}, they can provide accurate traffic predictions with the exploration of both temporal and spatial information. 

However, there are two significant shortcomings in existing traffic flow forecasting methods, which induce suboptimal performance.
\romannum{1}) \textit{Unable to capture dynamic non-pairwise relationships.} Existing methods often utilize graphs to characterize relations in the dynamic traffic system, which can merely capture pairwise relations. However, there could be abundant non-pairwise structural relations in the system. As is shown in Fig. \ref{fig:motivation}, a car accident could influence a range of locations in the dynamic traffic network, \R{and the model needs to capture this dynamic and non-pairwise impact. Similarly, locations around the residential area or the business area could share similar traffic conditions, and pairwise relationship modeling is inefficient when it comes to multiple nodes sharing similar properties.}
Therefore, the incapability of relation description limits the performance of spatio-temporal GNNs. 
(2) \textit{Unable to adequately explore high-order relationships.} 
Current methods often utilize GNNs to extract spatial features at each time step and aggregate the features across each time step using RNNs or TCNs~\cite{2020GMAN, 2020MTGNN,li2022dynamic}. These methods usually follow the paradigm of message passing to aggregate neighborhood information linearly. \R{Actually, real traffic data is complex and there could be abundant high-order information in the neighborhood of each observation}. The failure of modeling high-order spatio-temporal correlations hinders the model's capacity to make accurate traffic predictions.

To tackle the aforementioned drawbacks, we propose a novel method named \underline{Dy}namic \underline{H}ypergraph \underline{S}tructure \underline{L}earning (\method{}) for traffic flow prediction. To begin with, we extend prior road information into temporal graphs containing both spatial and temporal edges, which facilitates the exploration of spatio-temporal correlations using graph convolution. To model dynamic non-pairwise relations, we propose a Dynamic Hypergraph Structure Learning (DHSL) block that constructs a hypergraph among observations in the spatio-temporal network. To reduce the model parameters, the incidence matrix of the hypergraph is deduced from the node state representations at every timestamp. Then, a hypergraph convolution paradigm is proposed to update node representations using information from their associated hyperedges, which captures more complex relationships in the traffic network. Additionally, to explore high-order spatio-temporal relationships in the road network, we introduce an Interactive Graph Convolution (IGC) block, which explores neighborhood interaction using both combination and aggregation operators. The neighborhood interaction vector is then combined with the neighborhood embedding from linear aggregation to update node representations. 
In the end, we integrate these two blocks into a Multi-scale Holistic Correlation Extraction (MHCE) module, which first conducts temporal pooling with different granularity to model temporal patterns at different scales, Then, the data is fed into the two blocks, i.e., DHSL block and IGC block in parallel and their outputs are aggregated to update state representations iteratively.
Extensive experiments on three popular traffic datasets demonstrate that our proposed \method{} is capable of achieving superior performance compared with various state-of-the-art methods in different settings. In summary, the contributions of this paper are three-fold:
\begin{itemize}
    \item We propose a novel model named \method{} for traffic flow prediction, which models dynamic non-pairwise relationships using hypergraph structure learning and then conducts hypergraph convolution to capture more complex relationships in the traffic network.
    \item To explore high-order spatio-temporal relationships in the road network, \method{} introduces an interactive graph convolution block where node embeddings in the neighborhood are aggregated non-linearly.
    \item Comprehensive experiments are conducted on four well-known datasets and the results demonstrate that \method{} consistently outperforms various competing baselines.
\end{itemize}

\section{Related Work}\label{sec:related}

\subsection{Graph Neural Networks}
Graph neural networks (GNNs) have emerged as an effective tool which extend the deep neural networks to handle structured data~\cite{kipf2016semi,velivckovic2018graph,fang2022polarized,luo2022dualgraph,hu2021graphair,ju2022kernel,luo2022clear}, and have been extensively employed in a variety of applications including graph classification~\cite{ju2022ghnn}, node classification~\cite{zhu2020bilinear} and link prediction~\cite{zhang2018link}. Generally, existing GNN methods inherently follow an iterative message-passing paradigm \cite{gilmer2017neural} which recursively converts graphs into a low-dimensional embedding space to capture the structural information and node attributes. Recently, a range of GNN variants have been proposed to better extract spatial relationships among the structured data. 
For example, Bilinear GNN~\cite{zhu2020bilinear} attempts to model the interactions between neighboring nodes during message passing to enhance its representation capacity. SimP-GCN~\cite{jin2021node} seeks to preserve node similarity with sufficient exploration of the graph structure. 
\R{HGCN~\cite{yang2021hierarchical} uses graph capsules to obtain the hierarchical semantics. The differences between our DyHSL and HGCN lie in three points: (\romannum{1}) Our DyHSL focuses on dynamic graph learning but HGCN tackles the static graph. (\romannum{2}) HGCN builds disentangled graph capsules by underlying heterogeneous factors. In contrast, our DyHSL learns temporal hypergraphs via low-rank matrix decomposition, which is efficient and can simultaneously capture complicated spatial and temporal relationships for effective traffic flow forecasting. (\romannum{3}) Our method utilizes hypergraphs to model complex high-order traffic relationships while HGCN utilizes graph capsules to obtain the hierarchical semantics.}

\subsection{Hypergraph Neural Networks}

As a generalized form of graphs, a hypergraph consists of a collection of nodes and hyperedges~\cite{gao2020hypergraph,yang2022multi}. Unlike graph-structured data, hypergraphs can describe non-pairwise connections due to the fact that each hyperedge can link to many nodes. Due to the rising number of complex structured data in a variety of applications, such as recommender systems~\cite{xia2022hypergraph}, link prediction~\cite{li2013link}, and community detection~\cite{chien2018community}, hypergraph learning has recently attracted more attention. Since hypergraphs are a generalized form of graphs, these approaches are often an extension of graph neural networks. Using the concept of p-Laplacians~\cite{li2018submodular}, the early work aims to extend spectral methods on graphs to hypergraphs. Hypergraph neural network (HGNN)~\cite{feng2019hypergraph} is the first spatial approach on hypergraph learning that can discover latent node representations via the study of high-order structural information. However, the majority of these works concentrate on static hypergraphs. 

Recent efforts have been undertaken to learn from dynamic hypergraphs to address this issue. Dynamic hypergraph neural networks (DHGNN)~\cite{jiang2019dynamic} is the first effort to tackle the development of hyperedges; it builds dynamic hypergraphs and iteratively performs hypergraph convolution. \R{Compared to DHGNN that builds hypergraphs using kNN and K-Means algorithm to cluster node features, our \method{} explicitly learns the structure of the hypergraph based on low rank matrix decomposition, which is more efficient and effective.} Dynamic hypergraph convolutional network (DyHCN)~\cite{yin2022dynamic} provides spatio-temporal hypergraph convolution that investigates high-order correlations using the attention mechanism in the dynamic hypergraph. However, current studies usually focus on learning from hypergraphs, while our model use hypergraph structural learning to uncover dynamically complicated correlations in the traffic network.

\subsection{Traffic Flow Forecasting}
In recent years, traffic flow forecasting has received a surge of interest and a number of spatio-temporal forecasting methods have been proposed to solve this problem~\cite{zhao2019deep,cirstea2022towards,cirstea2022triformer,li2022spatial,liang2018deep}. The bulk of solutions to this problem is based on machine learning algorithms, which predict future traffic conditions based on spatio-temporal data gathered from numerous sensors. Traditional methods include k-nearest neighbors algorithm (kNN)~\cite{luo2019spatiotemporal}, autoregressive integrated moving average~\cite{box2015arima} (ARIMA) and support vector machines~\cite{sun2014traffic} (SVM) are often incapable of sufficiently modeling spatial relationships. With the development of deep neural networks, deep learning-based methods have become the mainstream solutions. The essence of these methods is to model the spatio-temporal correlations in the traffic data using deep neural networks. Among various neural network architectures, graph neural networks (GNNs) are well-suited for extracting structured spatial relationships in road networks while sequential neural networks can easily extract temporal relationships~\cite{li2022dynamic}. 

Recently, various spatio-temporal GNN methods have been proposed, capturing complicated spatial and temporal relationships for effective traffic prediction~\cite{2020GMAN, 2020MTGNN,li2022dynamic, 2020STSGCN, 2019ASTGCN}. 
\R{For instance, STSGCN~\cite{2020STSGCN} constructs a spatio-temporal graph and perform localized graph convolution on it. Compared to STSGCN which focuses on capturing pairwise and localized spatio-temporal dependencies, the proposed \method{} learns the hypergraph structure underlying the traffic data, which helps the model to capture long-range and non-pairwise relations. ASTGCN~\cite{2019ASTGCN} incorporates the attention mechanism into spatio-temporal graphs. However, this causes quadratic complexity, whereas the proposed \method{} achieves linear complexity with respect to both the size of the graph and the input length of observations (see \ref{sec:complexity}). LRGCN~\cite{li2019predicting} is also proposed to encode the spatio-temporal graph more efficiently. However, it tends to capture local and pairwise relationships both spatially and temporally. Compared to LRGCN, the proposed \method{} can capture longer dependencies among multiple nodes with dynamic hypergraph structure learning.}

\R{Hypergraphs have also been used in spatio-temporal forecasting~\cite{yi2020hypergraph, wang2021metro}. However, most existing works (\emph{e.g.} HGC-RNN~\cite{yi2020hypergraph} and DSTHGCN~\cite{wang2021metro}) require a predefined hypergraph as input, whereas our \method{} learns the structure underlying the spatio-temporal data.}
More detailed descriptions of these spatio-temporal GNN methods can be found in Section~\ref{baselines}.

\begin{table}
  \caption{Summary of notations and descriptions.}
  \label{symbol}
  \centering
  \begin{tabular}{p{0.1\textwidth}<{\centering} p{0.32\textwidth}<{\centering}}
    \toprule\midrule
     {Notations}&{Descriptions}\\\midrule
      $G=(V,E,\bm{A})$ &  Graph  \\ \midrule
 T  &  The length of observations  \\ \midrule
  $\bm{X}$ &  Graph signal tensor  \\ \midrule
 $\mathcal{G}=({\mathcal{V}},{\mathcal{E}})$ &  Hypergraph  \\ \midrule
   $\bm{\Lambda}$ &  Incidence matrix  \\ \midrule
 $G^H$ &  Temporal graph  \\ \midrule
  $\bm{h}_{i,(l)}^t$ &  State representation in prior graph convolution  \\ \midrule
   $\hat{\bm{A}}$ &  Adjacent matrix of the temporal graph  \\ \midrule
    $\bar{\bm{A}}$ &  Normalized adjacent matrix of the temporal graph \\ \midrule
     $\bm{F}$ &  Node embedding matrix in the first block  \\ \midrule
        $\bm{R}$ &  Node embedding matrix in the second block  \\ \midrule
       $\bm{{\Delta}}_l^\epsilon$ & The holistic state representation matrix \\ \midrule
        $\bm{\gamma}_i$ &  Final global embedding  \\\midrule\bottomrule
  \end{tabular}
\end{table}

\begin{figure*}[t]
\centering
\setlength{\belowcaptionskip}{-0.4cm}
\includegraphics[width=\textwidth,keepaspectratio=true]{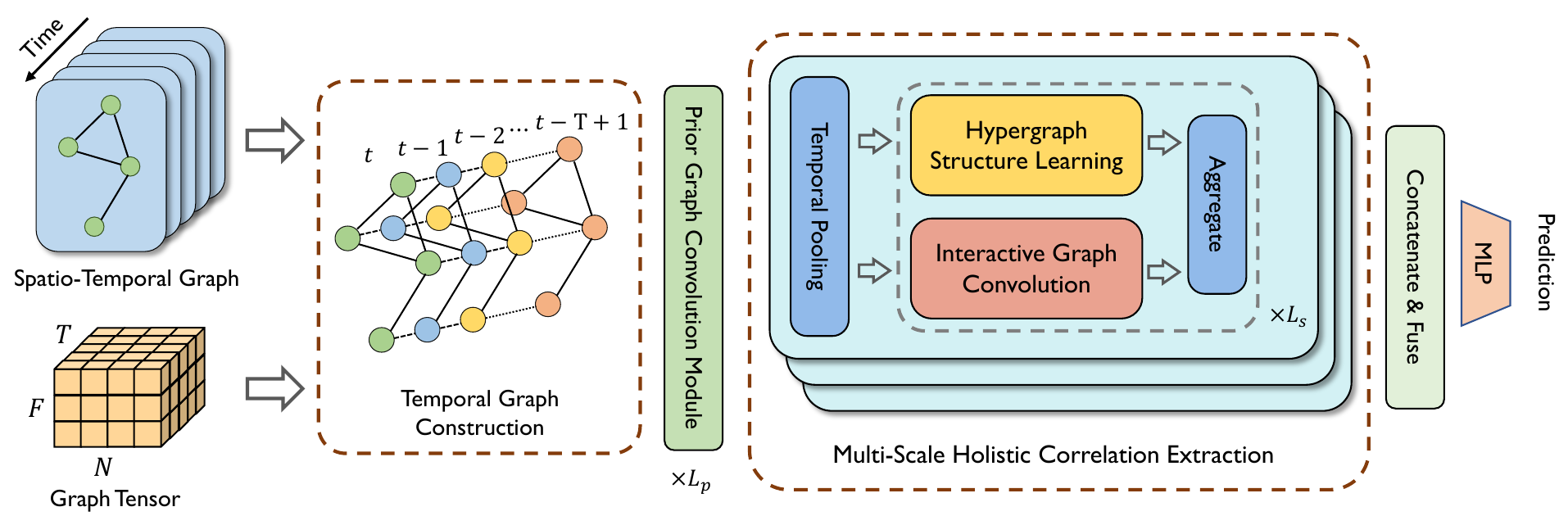}
\caption{Overview of our proposed \method{}. \method{} first generates temporal graphs based on the road network for prior graph convolution. In the multi-scale structured feature extraction module, we first utilize temporal pooling to aggregate data at different scales, and then feed hidden representations into both the hypergraph structure learning block and the neighborhood interaction block, followed by aggregation at every iteration. Finally, we fuse the output of different pooling sizes to output the final prediction.
}
\label{fig:framwork}
\end{figure*}
\section{Preliminaries}\label{sec:pre}
\subsection{Problem Definition}
In the problem of traffic flow forecasting, we are provided with a road network and historical traffic data. The road network is denoted as a weighted graph $G = (V,E,\bm{A})$ where $V$ denotes a set of $N$ nodes representing different locations in the road network. $E$ denotes a set of edges, which can be summarized in the weight adjacent matrix $\bm{A}\in \mathbb{R}^{N \times N}$. We characterize historical traffic observations into a graph signal tensor $\bm{X}=[\bm{X}_1,\bm{X}_2,\cdots,\bm{X}_T]\in \mathbb{R}^{T\times N\times F}$ where $T$ denotes the length of observations and $F$ denotes the dimension of node attributes. We aim to learn a function that maps the historical $T$ observations to predict the next $T'$ step traffic conditions. In formulation, 
\begin{equation}
\left[X_{t^0-T+1}, \cdots, X_{t^0} ; G\right] \stackrel{}{\longrightarrow}\left[\hat{X}_{t^0+1}, \hat{X}_{t^0+2}, \cdots, \hat{X}_{t^0+T^{\prime}}\right].
\end{equation}
Besides, the notations used in this paper are shown in Table \ref{symbol} for clarity.

\subsection{Hypergraph}

A hypergraph can be denoted as $\mathcal{G}=({\mathcal{V}},{\mathcal{E}})$ where ${\mathcal{V}}$ denotes the node set and ${\mathcal{E}}$ denotes the hyperedge set. Different from a graph, a hypergraph allows multiple nodes to be connected with a hyperedge. Similarly, a node can be associated with multiple hyperedges. Hence, we utilize a incidence matrix $\bm{\Lambda} \in \mathbb{R}^{|{\mathcal{V}}|\times |{\mathcal{E}}|}$ to characterize the structure of a hypergraph. Formally, for $v\in {\mathcal{V}}$ and $e \in {\mathcal{E}}$, we have:
\begin{equation}
\bm{\Lambda}(v,e)=\left\{
\begin{aligned}
1, \quad  & \text{if}\quad v\in e, \\
0, \quad  & \text{otherwise}. 
\end{aligned}
\right.
\end{equation}

We can simply extend the incidence matrix into a weighed form, 
\begin{equation}
\bm{\Lambda}(v,e)=\left\{
\begin{aligned}
w(v,e), \quad  & \text{if}\quad v\in e, \\
0, \quad  & \text{otherwise}, 
\end{aligned}
\right.
\end{equation}
where $w(v,e)$ denotes the interaction score of node $v$ and hyperedge $e$. 

\section{Methodology}\label{sec:method}
This work proposes a novel model named \method{} for traffic flow forecasting. Our \method{} first extends road information into temporal graphs containing both spatial and temporal edges, followed by graph convolution. To model dynamic non-pairwise relationships, we introduce a Dynamic Hypergraph Structure Learning (DHSL) block to construct a temporal hypergraph where nodes are all observations at all timestamps. To reduce the model parameters, the temporal hypergraph incidence matrix is deduced from each node state representation in a low-rank form. Then, we introduce hypergraph convolution to update node representations using information from their associated hyperedges. Moreover, we introduce an Interactive Graph Convolution (IGC) block to explore high-order spatio-temporal relationships in the road network. In this block, the neighborhood interaction is measured by both combination and aggregation operators to update node representations. Finally, we integrate these two blocks into a holistic multi-scale correlation extraction module, which first conducts temporal local pooling with different scales to model different temporal patterns, and then feeds data into two blocks parallelly. The framework of the model is illustrated in Fig. \ref{fig:framwork}.

\subsection{Prior Graph Encoder}
To begin, we utilize a prior graph encoder to capture basic spatio-temporal information from the road network. In our encoder, we first build a temporal graph based on road networks and then conduct prior graph convolution for node state representations.

\noindent\textbf{Temporal Graph Construction.}
Previous methods usually perform graph convolutional operation at every time step on road networks (or other pre-defined graphs)~\cite{fang2021spatial,li2022dynamic} to learn spatial correlations. However, they are unable to receive temporal information from other time steps during graph convolution, failing to model spatio-temporal interactions simultaneously. To tackle this issue, we introduce temporal graphs where nodes are observations determined by time-location pairs and are connected by both temporal edges and spatial edges. In this manner, spatio-temporal relationships could be modeled jointly during graph convolution. 

In particular, $T$ time steps yield $TN$ nodes $\{v_t\}_{t\in[1:T], v\in V}$ totally in the temporal graph ${G}^H$. At each time step, the spatial edges are identical to those in the original road network whereas the temporal edges exist when two observations are consecutive. In formulation, the adjacent matrix $\hat{\bm{A}} \in \mathcal{R}^{TN \times TN}$ with self-loop is derived as:
\begin{equation}\label{eq:graph}
\hat{\bm{A}}(v_{i}^{t},v_j^{t'})=\left\{\begin{array}{ll} \bm{A}_{ij} & t=t', \\ 1 & i=j, t'=t+1 \text{ or } t,\\ 0 & \text {otherwise}.\end{array}\right.
\end{equation}

\noindent\textbf{Prior Graph Convolution.}
Then, we present our graph convolution which learns joint spatio-temporal relationships in traffic data. Considering that temporal information has been inserted into temporal graphs, we utilize a standard graph convolution layer obeying the message passing mechanism~\cite{kipf2016semi} here. In detail, for each observation, we update its representation by aggregating the state embedding vectors of all its neighbors including itself at the previous layer. In formulation, the state representation of $v_i^t$ at the $l$-th layer $\bm{h}_{i,(l)}^{t}$ is calculated as follows:
\begin{equation}\label{eq:prior_graph_convolution}
\bm{h}_{i,(l)}^{t}=\phi(\sum_{v_j^{t'} \in \mathcal{N}(v_i^t)} \bar{\bm{A}}(v_{i}^{t},v_j^{t'}) \bm{f}_{j,(l-1)}^{t'} \bm{W}),
\end{equation}
where $\bar{\bm A}(v_{i}^{t},v_j^{t'})$ denotes the normalized adjacency matrix with $\sum_{j,t'} \bar{\bm{A}}(v_{i}^{t},v_j^{t'}) = 1$. $\phi(\cdot)$ is a non-linear activation function. 
\R{$\bm{f}_{j,(l-1)}^{t'}$ denotes the feature of node $j$ at time $t'$ and layer $l-1$, which is initially constructed by adding both spatial and temporal embeddings (representing the location and the time of a node respectively) to the original feature of the traffic network.}
After $L_p$-layer prior graph convolution, we can obtain the hidden state embedding of node $i$ at the $t$-th time step $\bm{h}_i^t = \bm{h}_{i,(L_p)}^{t} $.

\begin{figure}[ht]
\centering
\setlength{\belowcaptionskip}{-0.4cm}
\includegraphics[width=0.48\textwidth,keepaspectratio=true]{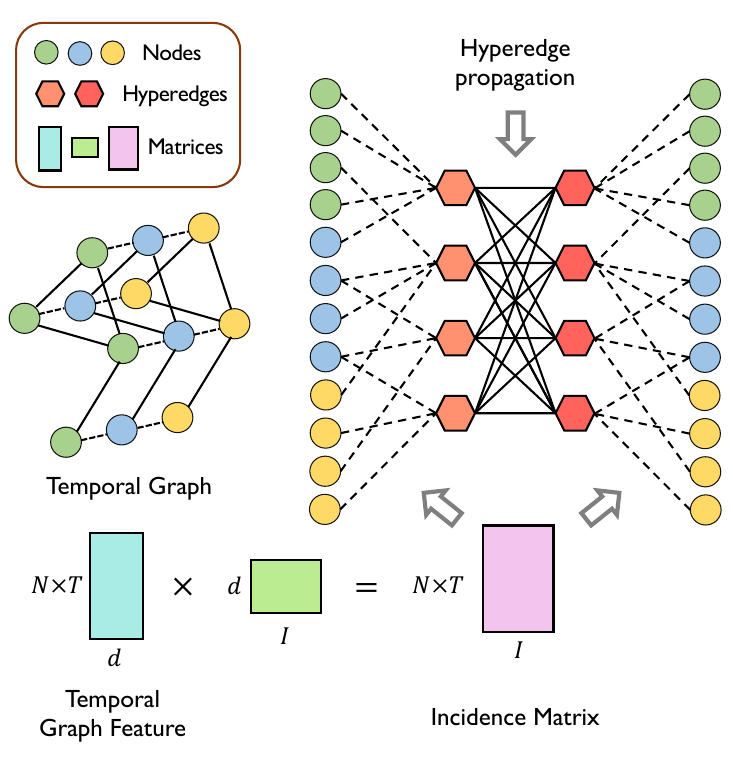}
\caption{The Dynamic Hypergraph Structure Learning (DHSL) block in \method{}. We first utilize node features to generate a low-rank incidence matrix. The hypergraph convolution operation first fuses information from connected nodes into hyperedge representations and then reconstructs node representation using associated hyperedge representations. 
}
\label{fig:hypergraph}
\end{figure}
\subsection{Dynamic Hypergraph Structure Learning}
Nevertheless, traffic systems are influenced by the road network and dynamic traffic conditions in tandem. Consequently, we need to model the real-time traffic situation based on hidden embeddings. 
Previous methods are mostly based on dynamic graphs/hypergraphs, which \R{have drawbacks like requiring predefined structues~\cite{yi2020hypergraph, wang2021metro}, high computation cost~\cite{2019ASTGCN, yang2021hierarchical} or failing to capture non-pairwise relations~\cite{2020STSGCN, li2019predicting}.} To tackle this, we turn to temporal hypergraphs for complex relation modeling where each node is also an observation at a given timestamp. \R{Moreover, we explicitly learn the hypergraph structure in a low-rank manner. }

\noindent\textbf{Temporal Hypergraph \R{Structure Learning}.} To explore dynamic complex relationships in traffic networks, we introduce learnable hypergraph structure matrices, which are optimized jointly with network parameters. To decrease the parameters and release overfitting, we seek to utilize matrix decomposition to construct a low-rank structure matrix.

In detail, the incidence matrix of the temporal hypergraph is formalized as $\bm{\Lambda} \in \mathbb{R}^{NT \times I}$ where $I$ is the number of the hyperedge. We decompose the matrix into two low-rank matrices using their hidden state representations as follows:
\begin{equation}
\label{eq:incidence}
\bm{\Lambda}=\mathbf{H} \mathbf{W}
\end{equation}
where $\mathbf{H} \in \mathbb{R}^{NT\times d}$ is derived by stacking all state representations and $\mathbf{W} \in \mathbb{R}^{d\times I}$ is learnable weight matrix. In this way, learning the incidence matrix merely introduce $\mathcal{O}(I\times d)$ parameters ($d<<NT$), which can significantly improve the model efficiency. 

\noindent\textbf{Temporal Hypergraph Convolution.}
Then, we introduce a hypergraph convolution paradigm for learning from the temporal hypergraph, which can extract high-order complex information from dynamic traffic networks. To be specific, at every layer, we first generate each hyperedge embedding by aggregating information from all its connected nodes. Then, hypergraph embeddings are used to update node embedding for high-order correlation learning in the traffic network. The overall process is summarized in Fig. \ref{fig:hypergraph}. 

In the matrix form, the hyperedge embedding matrix  $\bm{E} \in \mathbb{R}^{I \times d}$ is derived from the state representation matrix and the incidence matrix:
\begin{equation}
    \bm{E} = \phi(\bm{U}\bm{\Lambda}^T \bm{H})+\bm{\Lambda}^T \bm{H}.
\end{equation}
where we additionally introduce a learnable matrix $\bm{U}_H \in \mathbb{R}^{I\times I}$ to characterize the implicit relations among hyperedges.
Then, these hyperedge embeddings are aggregated to generate the node embedding matrix:
\begin{equation}\label{eq:message}
    \mathbf{F} = \bm{\Lambda} \bm{E}=\bm{\Lambda}(\phi(\bm{U}\bm{\Lambda}^T \bm{H})+\bm{\Lambda}^T \bm{H}).
\end{equation}
In this way, we can learn complex non-pairwise correlation in the traffic network at each timestamp by stacking $L_H$ hypergraph convolution layers, which outputs the updated node embedding matrix $ \mathbf{F} = BLOCK_H(\mathbf{H}) \in\mathbb{R}^{NT\times d}$.

\begin{figure}[t]
\centering
\setlength{\belowcaptionskip}{-0.4cm}
\includegraphics[width=0.45\textwidth,keepaspectratio=true]{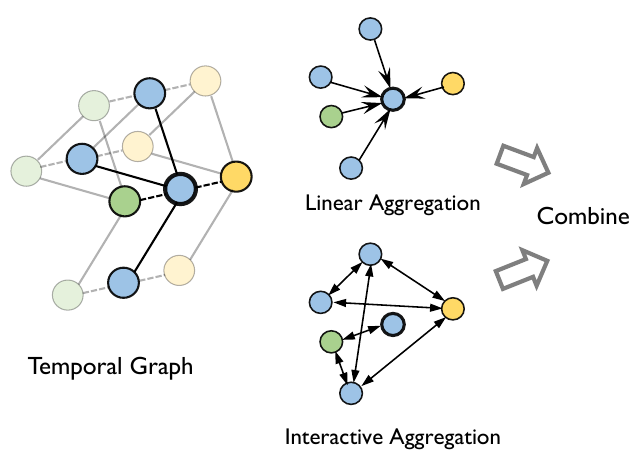}
\caption{The Interactive Graph Convolution (IGC) block in \method{}. During interactive aggregation, we first model the interaction of neighboring node pairs and then aggregate all these pairs into interactive aggregation. Moreover, we utilize standard linear aggregation and concatenate the outputs from both lines. 
}
\label{fig:interactive}
\end{figure}
\subsection{Interactive Graph Convolution}

Additionally, there is still abundant high-order spatio-temporal correlation in the road network, which cannot be captured by our prior graph convolution. Hence, we introduce another block that utilizes interactive neighborhood aggregation which fully models high-order information in the temporal graph using both combination and aggregation operators. In this way, we can learn the signals under the co-occurrence of neighbors.

In detail, we recall the temporal graph $G^H$ but further model the interaction of $v_{j}^{t'}$ and $v_{j'}^{t''}$ from $\mathcal{N}(v_{i}^{t})$. Formally, given each state representation $\bm{h}_{i}^t$, its interactive representation vector is written as follows:
\begin{equation}
    \bm{\pi}_{i}^t = \operatorname{AGG}(\{\operatorname{COM}(\bm{h}_{j}^{t'},\bm{h}_{j'}^{t''}), \forall {v_j^{t'},v_{j'}^{t''} \in \mathcal{N}(v_i^t)} \}),
\end{equation}
where $\operatorname{COM}$ calculates the coupled representation for each node pair and $\operatorname{AGG}$ aggregates all neighborhood node pairs for each central node, which needs to be permutation-invariant. In this work, we utilize separate projectors followed by Hadamard produce to implement $\operatorname{COM}$ and sum-pooling followed by an activation function to implement $\operatorname{AGG}$:
\begin{equation}\label{eq:sec}
    \bm{\pi}_{i}^t=\phi(\sum_{v_j^{t'},v_{j'}^{t''} \in \mathcal{N}(v_i^t)}  \bar{\bm{A}}_{i_{t},j_{t'}} \bar{\bm{A}}_{i_{t},{j'}_{t''}} \bm{h}_{j}^{t'}\bm{W}_1\odot \bm{h}_{j'}^{t''} \bm{W}_2),
\end{equation}
where $\odot$ denotes the Hadamard product of two vectors, $\bm{W}_1 $ and $\bm{W}_2 $ denotes a learnable weight matrix. Eq.~\ref{eq:sec} can be rewritten as:
\begin{equation}
    \phi(\sum_{v_j^{t'} \in \mathcal{N}(v_i^t)} \bar{\bm{A}}_{i_{t},j_{t'}} \bm{h}_{j}^{t'}\bm{W}_1  \odot \sum_{v_j^{t'} \in \mathcal{N}(v_i^t)} \bar{A}_{i_{t},j_{t'}} \bm{h}_{j}^{t'}\bm{W}_2 ).
\end{equation}

Moreover, we also incorporate linear aggregation of the neighborhood into our block, which can also provide basic information about the road network. Hence, the updated temporal state representation $\bm{r}_{i}^{t}$ is formulated as follows:
\begin{equation}
    \bm{r}_{i}^{t} = \bm{\pi}_{i}^t+ \phi(\sum_{v_j^{t'} \in \mathcal{N}(v_i^t)} \bar{\bm{A}}_{i_{t},j_{t'}} \bm{h}_{j}^{t'}\bm{W}_3).
\end{equation}

Finally, we can derive the updated state representation matrix $\bm{R}=BLOCK_I(\bm{H})\in \mathbb{R}^{NT \times d}$ where each line of the matrix corresponds to each observation. The whole procedure of this block is illustrated in Fig.~\ref{fig:interactive}.

\subsection{Multi-scale Holistic Correlation Extraction}
\label{sec:complexity}
In this part, we integrate two introduced blocks into a holistic complex correlation extraction framework. These two blocks can play a complementary role since hypergraph structure learning tends to extract dynamic signals beyond pairwise relationships whereas interactive graph convolution tends to learn from high-order relationships on the basis of the road network. Motivated by diverse patterns in the traffic network, we first conduct local pooling on the embedding sequences using different window sizes and then extract correlation at different scales using the two blocks. 

In particular, after prior graph convolution,
we first determine a few candidates of window sizes. Taking the window size $\epsilon$ as an example, we generate a list of subsequence embeddings for each node using local-max pooling, $\{\bm{\delta}^1_i,\cdots, \bm{\delta}^{T/\epsilon}_i \} $, where $\bm{\delta}^k_i=Pool(\bm{h}^{k\epsilon-\epsilon+1}_i,\cdots,\bm{h}^{k\epsilon}_i)$. Then, we feed the concatenated subsequence embeddings $\bm{\Delta}^\epsilon \in \mathbb{R}^{NT/\epsilon \times d}$ into both the dynamic hypergraph structure learning block and the interactive graph convolution block. In the first block, we construct the temporal hypergraph for hypergraph convolution while in the second block we build temporal graphs along the subsequence for inter-graph convolution. Finally, we take the average of both outputs. Moreover, this procedure is conducted in an iterative manner. 
In formulation, we have the holistic state representation matrix $\bm{{\Delta}}^{\epsilon}_l \in \mathbb{R}^{NT/\epsilon \times d}$ at the $l$-th layer:
\begin{equation}
    \bm{{\Delta}}_{l}^\epsilon = \frac{1}{2} (BLOCK_H( \bm{{\Delta}}_{l-1}^\epsilon)+BLOCK_I( \bm{{\Delta}}_{l-1}^\epsilon))
\end{equation}
where $BLOCK_H(\cdot)$ and $BLOCK_I(\cdot)$ denote the DHSL and the IGC blocks. After stacking $L_s$ layers, we get $\bm{{\Delta}}_{L_s}^\epsilon$ eventually. Then we decompose the matrix into a tensor $\Gamma^\epsilon \in \mathbb{R}^{N\times T/\epsilon \times d}$ and then aggregate the embedding matrix along the time dimension, producing sequence embeddings $\bm{\gamma}_i^{\mu}\in \mathbb{R}^d$ for each node. In practice, mean-pooling is utilized for aggregation. Considering that various scales could characterize diverse intrinsic properties such as variances in traffic data, we choose $J$ different window sizes $\epsilon_1$, $\epsilon_2$, $\cdots, \epsilon_J$, which result in three granularity-aware sequence embeddings $\bm{\gamma}_i^{\epsilon_1}$, $\bm{\gamma}_i^{\epsilon_2}$,$\cdots$ $\bm{\gamma}_i^{\epsilon_J}$, respectively. 

In a nutshell, we obtain global sequence embeddings and then adaptively decide their contributions to the final global embeddings.
To be specific, we introduce $J$ learnable parameters $\{w^{\epsilon_{j}}\}_{j=1}^J$, and the final global embedding is written as:
\begin{equation}
    \bm{\gamma}_i =\frac{\sum_{j=1}^J\exp(w^{\epsilon_j})\bm{\gamma}_i^{\epsilon_j}}{\sum_{j=1}^J\exp(w^{\epsilon_j}) }.
\end{equation}

The global embedding $\bm{\gamma}_i$ is then concatenated with the local embedding vector at the last time step, i.e., $\bm{h}^T$, forming the final output $\bm{y}\in\mathbb{R}^{T'\times 1}$ through a fully-connected layer for each node. 
The model is optimized using the standard mean absolute error (MAE) loss for regression. The whole algorithm is summarized in Algorithm~\ref{alg1}. 

\begin{algorithm}[t]
\caption{Training Algorithm of \method{}}
\label{alg1}
\begin{algorithmic}[1]
\REQUIRE Road network ${G}$, Graph Signal Tensor $\bm{X}$;
\ENSURE Predictions of traffic signals in $T'$ future time steps;

\STATE Generate the temporal graph using Eq.~\ref{eq:graph};
\REPEAT
\STATE Compute each $\bm{h}_i^t$ using Eq.~\ref{eq:prior_graph_convolution};
\STATE Conduct temporal pooling;
\FOR{$i=1, \cdots,L_s$}
\STATE Calculate the output of the Dynamic Hypergraph Structure Learning (DHSL) block;
\STATE Calculate the output of the Interactive Graph Convolution (IGC) block;
\STATE Update state representations by taking the average; 
\ENDFOR
\STATE Calculate the final prediction and the MAE loss;
\STATE Update the model parameters using stochastic gradient descend (SGD);
\UNTIL convergence
\end{algorithmic}
\end{algorithm}

The computational complexity of Algorithm~\ref{alg1} mainly depends on Step 4-5. Recall that the number of nodes and the length of observations are denoted as $N$ and $T$, respectively. $L_s$ denotes the number of hidden layers in multi-scale holistic correlation extraction, respectively. $||A||_0$ denotes the number of nonzeros in the adjacency matrix. $d$ denotes the hidden dimension. On the one hand, the complexity of our dynamic hypergraph structure learning block is about $\mathcal{O}(NTIL_s)$.
On the other hand, the complexity of the interactive graph convolution block is about $\mathcal{O}(L_s||A||_0FT+L_sNF^2T+NTF)$, \R{which is linearly related to both $||A||_0$ and $T$. This demonstrates that the computation time grows linearly with the size of the traffic network ($||A||_0$) and length of observation ($T$).} Moreover, our method adopts a low-rank incidence matrix, which is more efficient with fewer parameters.

\section{Experiments}\label{experiment}
In this section, we conduct experiments to demonstrate the effectiveness of the proposed \method{}. Moreover, we provide comprehensive analysis on the performance of the model as well as the role of dynamic hypergraph structure learning.
\subsection{Experimental Setup}
\subsubsection{Datasets}
{
\begin{table}
    \centering
    \caption{The summary of the datasets used in the experiments.}
    \label{tab:dataset}
    \small
    \begin{tabular}{ccccc}
    \toprule\midrule
        \textbf{Dataset}     & $|{V}|$  & $| E|$ & \textbf{Time Steps} & \textbf{Time Range} \\ \midrule
        PEMS03      & 358 &  547    & 26,208    & 09/2018 - 11/2018\\ 
        PEMS04      & 307 &  340    & 16,992    & 01/2018 - 02/2018\\ 
        PEMS07      & 883 &  866    & 28,224    & 05/2017 - 08/2017\\ 
        PEMS08      & 170 &  295    & 17,856    & 07/2016 - 08/2016\\ 
    \midrule\bottomrule
    \end{tabular}
\end{table}
}
In the experiments, we use four widely adopted, publicly available datasets: PEMS03, PEMS04, PEMS07 and PEMS08. These datasets are collected by California Transportation Agencies (CalTrans) Performance Measurement Systems (PEMS)\footnote{https://pems.dot.ca.gov/}. The traffic data are collected every 30 seconds and aggregated into 5-minute time steps. We use the standard data processed and released by~\cite{2020STSGCN} in alignment with previous works.

The spatial graph for each dataset is constructed with regard to the real road network. In all the experiments, standard spatial graphs are adopted which are also provided by~\cite{2020STSGCN}. The detailed statistics about these spatial graphs and time ranges of the aforementioned datasets are listed in Table~\ref{tab:dataset}.

\subsubsection{Evaluation Setting and Metrics}
The model takes 60 minutes (12 time steps) of historical data as input and outputs the prediction of traffic flow in the next 60 minutes (12 time steps). We follow the standard dataset split and use 60\% of the data for training, 20\% for evaluation and the remaining 20\% for testing the model's performance. To evaluate the prediction error, standard metrics are used including Mean Absolute Error (MAE), Root Mean Squared Error (RMSE), and Mean Absolute Percentage Error (MAPE).

{
\begin{table*}[t]
    \centering
    \caption{Forecasting errors on PEMS03, PEMS04, PEMS07 and PEMS08 datasets.}
    \label{tab:main_exp}
    \begin{tabular}{c ccc c ccc c ccc c ccc}
        \toprule
        \midrule  
        \multirow{2}{*}{\textbf{Model}}  & \multicolumn{3}{c}{\textbf{PEMS03}}    && \multicolumn{3}{c}{\textbf{PEMS04}}      && \multicolumn{3}{c}{\textbf{PEMS07}}      && \multicolumn{3}{c}{\textbf{PEMS08}}\\\cmidrule{2-4} \cmidrule{6-8} \cmidrule{10-12} \cmidrule{14-16}
                                & MAE & RMSE & MAPE             && MAE & RMSE & MAPE               && MAE & RMSE & MAPE               && MAE & RMSE & MAPE               \\ \midrule
        HA                      & 31.58 & 52.39 & 33.78\%       && 38.03 & 59.24 & 27.88\%         && 45.12 & 65.64 & 24.51\%         && 34.86 & 59.24 & 27.88\%         \\ 
        ARIMA                   & 35.41 & 47.59 & 33.78\%       && 33.73 & 48.80 & 24.18\%         && 38.17 & 59.27 & 19.46\%         && 31.09 & 44.32 & 22.73\%\\ 
        VAR                     & 23.65 & 38.26 & 24.51\%       && 24.54 & 38.61 & 17.24\%         && 50.22 & 75.63 & 32.22\%         && 19.19 & 29.81 & 13.10\%         \\  
        SVR                     & 21.97 & 35.29 & 21.51\%       && 28.70 & 44.56 & 19.20\%         && 32.49 & 50.22 & 14.26\%         && 23.25 & 36.16 & 14.64\%         \\ 
        FC-LSTM                 & 21.33 & 35.11 & 23.33\%       && 26.77 & 40.65 & 18.23\%         && 29.98 & 45.94 & 13.20\%         && 23.09 & 35.17 & 14.99\%         \\ 
        TCN                     & 19.32 & 33.55 & 19.93\%       && 23.22 & 37.26 & 15.59\%         && 32.72 & 42.23 & 14.26\%         && 22.72 & 35.79 & 14.03\%         \\  
        TCN(w/o causal)         & 18.87 & 32.24 & 18.63\%       && 22.81 & 36.87 & 14.31\%         && 30.53 & 41.02 & 13.88\%         && 21.42 & 34.03 & 13.09\%         \\  
        GRU-ED                  & 19.12 & 32.85 & 19.31\%       && 23.68 & 39.27 & 16.44\%         && 27.66 & 43.49 & 12.20\%         && 22.00 & 36.22 & 13.33\%         \\       
        DSANet                  & 21.29 & 34.55 & 23.21\%       && 22.79 & 35.77 & 16.03\%         && 31.36 & 49.11 & 14.43\%         && 17.14 & 26.96 & 11.32\%         \\       
        STGCN                   & 17.55 & 30.42 & 17.34\%       && 21.16 & 34.89 & 13.83\%         && 25.33 & 39.34 & 11.21\%         && 17.50 & 27.09 & 11.29\%         \\ 
        DCRNN                   & 17.99 & 30.31 & 18.34\%       && 21.22 & 33.44 & 14.17\%         && 25.22 & 38.61 & 11.82\%         && 16.82 & 26.36 & 10.92\%         \\ 
        GraphWaveNet            & 19.12 & 32.77 & 18.89\%       && 24.89 & 39.66 & 17.29\%         && 26.39 & 41.50 & 11.97\%         && 18.28 & 30.05 & 12.15\%         \\ 
        ASTGCN(r)               & 17.34 & 29.56 & 17.21\%       && 22.93 & 35.22 & 16.56\%         && 24.01 & 37.87 & 10.73\%         && 18.25 & 28.06 & 11.64\%         \\ 
        STG2Seq                 & 19.03 & 29.83 & 21.55\%       && 25.20 & 38.48 & 18.77\%         && 32.77 & 47.16 & 20.16\%         && 20.17 & 30.71 & 17.32\%         \\ 
        \R{DHGNN}                & \R{16.99} & \R{28.16} & \R{17.02\%}       && \R{20.96} & \R{32.64} & \R{14.55\%}         && \R{22.73} & \R{35.67} & \R{10.27\%}  && \R{18.10} & \R{28.53} & \R{10.82\%}    \\ 
        \R{LRGCN}                 & \R{17.96} & \R{30.37} & \R{18.54\%}       && \R{20.02} & \R{33.21} & \R{17.43\%}         && \R{22.53} & \R{36.27} & \R{10.33\%}  && \R{16.41} & \R{26.37} & \R{11.88\%}    \\ 
        LSGCN                   & 17.94 & 29.85 & 16.98\%       && 21.53 & 33.86 & 13.18\%         && 27.31 & 41.46 & 11.98\%         && 17.73 & 26.76 & 11.20\%         \\      
        STSGCN                  & 17.48 & 29.21 & 16.78\%       && 21.19 & 33.65 & 13.90\%         && 24.26 & 39.03 & 10.21\%         && 17.13 & 26.80 & 10.96\%         \\      
        AGCRN                   & 15.98 & 28.25 & 15.23\%       && 19.83 & 32.26 & 12.97\%   && 22.37 & 36.55 &  9.12\%    && 15.95 & 25.22 & 10.09\%         \\
        \R{HGC-RNN}                 & \R{17.04} & \R{28.17} & \R{17.99\%}       && \R{20.39} & \R{32.42} & \R{13.58\%}         && \R{22.4} & \R{35.37} & \R{9.69\%}  && \R{16.28} & \R{25.6} & \R{10.68\%}    \\ 
        \R{DSTHGCN}                 & \R{17.09} & \R{27.54} & \R{16.14\%}       && \R{21.12} & \R{33.54} & \R{14.24\%}         && \R{23.22} & \R{36.93} & \R{9.95\%}  && \R{16.49} & \R{25.98} & \R{10.22\%}    \\ 
        STFGNN                  & 16.77 & 28.34 & 16.30\%       && 20.48 & 32.51 & 16.77\%         && 23.46 & 36.60 &  9.21\%         && 16.94 & 26.25 & 10.60\%         \\       
        STGODE                   & 16.50 & 27.84 & 16.69\%       && 20.84 & 32.82 & 13.77\%         && 22.59 & 37.54 & 10.14\% && 16.81 & 25.97 & 10.62\%         \\
        Z-GCNETs                & 16.64 & 28.15 & 16.39\%       && 19.50 & 31.61 & 12.78\%        && {21.77} & {35.17} & 9.25\%   && {15.76} & {25.11} & {10.01}\% \\
        {STG-NCDE}       & {15.57} & {27.09}    &   {15.06}\% && {19.21}    & {31.09}    &  {12.76}\%   && {20.53} & {33.84} & {8.80}\%    && {15.45} & {24.81} &  {9.92}\% \\
        {DSTAGNN}       & {15.57} & {27.21}    &   {14.68}\% && {19.30}    & {31.46}    &  {12.70}\%   && {21.42} & {34.51} & {9.01}\%    && {15.67} & {24.77} &  {9.94}\% \\\midrule
        \textbf{\method{}}  & \textbf{15.49} & \textbf{27.06} & \textbf{14.38}\%       && \textbf{17.66} & \textbf{29.46} & \textbf{12.42}\%      && \textbf{18.84} & \textbf{31.65} & \textbf{8.11}\%         && \textbf{14.01} & \textbf{22.91} & \textbf{8.60}\% \\
        \midrule
        \bottomrule
    \end{tabular}
\end{table*}

}
\subsubsection{Baselines}\label{baselines}
We compare the proposed \method{} with a wealth of baselines, ranging from traditional statistic-based methods to the recent neural network-based methods. The details about the baselines are listed as follows:

\textit{Traditional statistic-based methods:}
\begin{itemize}
    \item \textbf{HA}: 
    Historical Average uses weighted averages of historical data as predictions for future values.
    \item \textbf{ARIMA}~\cite{box2015arima}:
    Auto-Regressive Integrated Moving Average is a well-known statistic-based model widely used for time series forecasting.
    \item \textbf{VAR}~\cite{lutkepohl2005new}: 
    Vector Auto-Regression is another traditional method for time series forecasting.
    \item \textbf{SVR}: Support Vector Regression uses a support vector machine for regression.
\end{itemize}

\textit{Neural network methods without the spatial graph:}
\begin{itemize}
    \item \textbf{FC-LSTM}~\cite{FC-LSTM}: LSTM~\cite{hochreiter1997long} network with fully connected hidden units is a well-known model for capturing sequential dependencies.
    \item \textbf{TCN}~\cite{TCN2018}: Temporal Convolution Network uses a stack of dilated casual convolution layers with exponentially growing dilation factors. The performance without casual convolution is also presented.
    \item \textbf{GRU-ED}~\cite{2014GRU}: Gated Recurrent Units with encoder-decoder architecture is also a commonly used baseline for multi-step time series forecasting.
    \item \textbf{DSANet}~\cite{Huang2019DSANet}: Dual Self-Attention Network is a powerful multivariate time series forecasting method that utilizes both CNNs and the self-attention mechanism.
\end{itemize}

\textit{Neural network methods using the spatial graph:}
\begin{itemize}
    \item \textbf{STGCN}~\cite{bing2018stgcn}: This model combines graph convolutions with temporal convolutions.
    \item \textbf{DCRNN}~\cite{2017DCRNN}: Diffusion Convolutional Recurrent Neural Network uses diffusion convolution to replace the fully-connected layers in GRU~\cite{2014GRU}.
    \item \textbf{Graph WaveNet}~\cite{GWNet}: This method uses dilated convolution and diffusion graph convolution, and also proposes a self-adaptive adjacency matrix.
    \item \textbf{ASTGCN}~\cite{2019ASTGCN}: This model adopts both spatial attention and temporal attention. The (r) notation in Table~\ref{tab:main_exp} indicates that only recent components of modeling periodicity are included for a fair comparison.
    \item \textbf{STG2Seq}~\cite{bai2019STG2Seq}: This work adopt a sequence-to-sequence framework with multiple gated graph convolution module and the attention mechanism for multi-step prediction.
    \R{\item \textbf{DHGNN}~\cite{jiang2019dynamic}: This method uses kNN and K-means algorithms to learn hypergraphs and perform convolutions based on them. We adapt it in traffic flow forecasting.}
    \R{\item \textbf{LRGCN}~\cite{li2019predicting}: This method uses LSTM~\cite{hochreiter1997long} and grapn convolution networks to encode spatio-temporal graphs. We also adapt this method in traffic flow forecasting.}
    \item \textbf{LSGCN}~\cite{huang2020lsgcn}:This work uses spatial gated blocks and gated linear units in conjunction with the attention mechanism and graph convolution.
    \item \textbf{STSGCN}~\cite{2020STSGCN}: This method models the spatial and temporal dependencies synchronously and adopts local spatio-temporal subgraph modules.
    \item \textbf{AGCRN}~\cite{bai2020AGCRN}: Adaptive Graph Convolutional Recurrent Network learns the node-specific features and the hidden inter-dependencies among nodes.
    \R{\item \textbf{HGC-RNN}~\cite{yi2020hypergraph}: This method uses hypergraph convolution in conjunction with RNNs. We adapt this method in traffic flow forecasting.}
    \R{\item \textbf{DSTHGCN}~\cite{wang2021metro}: This method uses dynamic hypergraphs to process spatio-temporal data. Since it is originally proposed for metro passenger flow prediction, we adapt this method for traffic flow forecasting.}
    \item \textbf{STFGNN}~\cite{li2021stfgnn}: This model utilizes a spatial fusion graph and a generated temporal graph.
    \item \textbf{STGODE}~\cite{fang2021STODE}: This model adopts ordinary differential equations (ODEs) for traffic flow forecasting.
    \item \textbf{Z-GCNETs}~\cite{chen2021ZGCNET}: This method introduces the zigzag persistence that can be used to track important topological features from the observed data over time.
    \item \textbf{STG-NCDE}~\cite{choi2022stgncde}: This work designs two neural controlled differential equations (NECDs) for spatial and temporal processing, respectively.
    \item \textbf{DSTAGNN}~\cite{lan2022dstagnn}: This model learns a spatio-temporal graph and uses multi-head attention to represent dynamic spatial relevance.
\end{itemize}

\subsubsection{Implementation Details}
The proposed model is implemented with PyTorch and is trained on an NVIDIA RTX GPU. In the Prior Graph Convolution, we use 6 convolutional layers (\emph{i.e.} $L_p=6$). In the Dynamic Hypergraph Structure Learning block, we use 32 hyperedges (\emph{i.e.} $I=32$). In Multi-scale Holistic Correlation Extraction, we use a set of 6 window sizes, \emph{i.e.} $J=6$ and $\epsilon\in \{1, 2, 3, 4, 6, 12\}$, and the number of hidden layers is set to 2 (\emph{i.e.} $L_s=2$). The dimension for the hidden features is set to $64$ (\emph{i.e.} $d=64$). More analysis about hyperparameters in our model can be found in Section \ref{sec:hyper}. For optimization, we use Adam optimizer~\cite{kingma2014adam} and train the model for 100 epochs, setting the learning rate to 0.001 and batch size to 32.

\subsection{The Performance of \method{} }

The performance of \method{} in comparison with baselines are listed in Table~\ref{tab:main_exp}. According to the results, we have several observations described as follows.

Firstly, the proposed \method{} achieves a consistent lead in all three metrics on four datasets, which shows the superiority of the proposed Dynamic Hypergraph Structure Learning (DHSL) block, Interactive Graph Convolution (IGC) block and the Multi-scale Holistic Correlation Extraction (MHCE). More specifically, the significant improvement can be attributed to three aspects: \romannum{1}) The model is able to learn complex dynamic structures among multiple nodes. Most existing works (\emph{e.g.} DCRNN~\cite{2017DCRNN}, STSGCN~\cite{2020STSGCN}, DSTAGNN~\cite{lan2022dstagnn}) models \emph{pair-wise} dependencies between nodes, whereas our model also considers the correlations among \emph{multiple} nodes. \romannum{2}) The proposed IGC block is able to capture higher-order interactions in the graph, while many previous works fall short in learning the interaction of adjacent nodes. \romannum{3}) The multi-scale framework is suitable for capturing the traffic pattern with different periodicity, whereas many existing algorithms ignore this. 
    
Besides, the proposed model achieves a greater improvement (\emph{i.e.} \textbf{8.2}\% relative improvement on MAE, \textbf{6.5}\% relative improvement on RMSE and \textbf{7.8}\% relative improvement on MAPE) on the PEMS07 dataset, which is the largest dataset among the four. This suggests that our proposed model is able to handle large-scale traffic data more efficiently.

Generally, traditional statistic-based methods (\emph{e.g.} ARIMA~\cite{box2015arima}, SVR) perform worse than the deep learning algorithms, because they require the stationary assumption (which is often violated in the real world) and ignore the spatial topology. Some deep learning methods (\emph{e.g.} FC-LSTM~\cite{FC-LSTM}, TCN~\cite{TCN2018}, GRU-ED~\cite{2014GRU}) only consider the temporal dependencies and thus fall short in capturing spatial correlations, which result in weak performance compared to GNN-based methods. The recently proposed GNN-based algorithms are able to capture both spatial and temporal relations, which leads to better performances compared to the previous methods. Among these methods, \method{} achieves the best performance, which shows the effectiveness of the proposed DHSL block, IGC block and MHCE.

\subsection{Scalability Studies}

{
\begin{table}[ht]
    \centering
    \caption{\R{The number of parameters, training and testing time (measured in terms of seconds) of \method{} compared to some baseline models. The results show that our model has less parameters and comparable training/testing time with superior performance.}}
    \begin{tabular}{cccc}
        \toprule
        \midrule
         \R{Model} & \R{\# Parameters} & \R{Training Time (/epoch)} & \R{Testing Time}
         \\\midrule
        \R{STGODE} & \R{714K} & \R{92.49} & \R{8.5} \\
        \R{DSTAGNN} & \R{3.58M} & \R{190.5} & \R{15.8} \\
        \R{\method{}} & \R{256K} & \R{104.5} & \R{14.2} \\
         \midrule
         \bottomrule
    \end{tabular}
    \label{tab:efficiency}
\end{table}
}
\R{
In addition to the prediction performance, the computational efficiency is also a crucial factor in real-world practice. In this part, we compare the number of parameters, training and testing times of our \method{} compared with two baselines.
As can be seen in Table~\ref{tab:efficiency}, our model has the least parameters among the three models. Additionally, the training and testing time of our model is comparable with current SOTA methods. As with most traffic flow forecasting models, the proposed \method{} scales linearly with respect to the size of the graph and the length of observations (explained in Section~\ref{sec:complexity}). In conclusion, the proposed \method{} is also competitive in terms of the number of parameters and training/testing time.
}

\begin{figure*}[ht]
\centering
\includegraphics[width=0.95\textwidth,keepaspectratio=true]{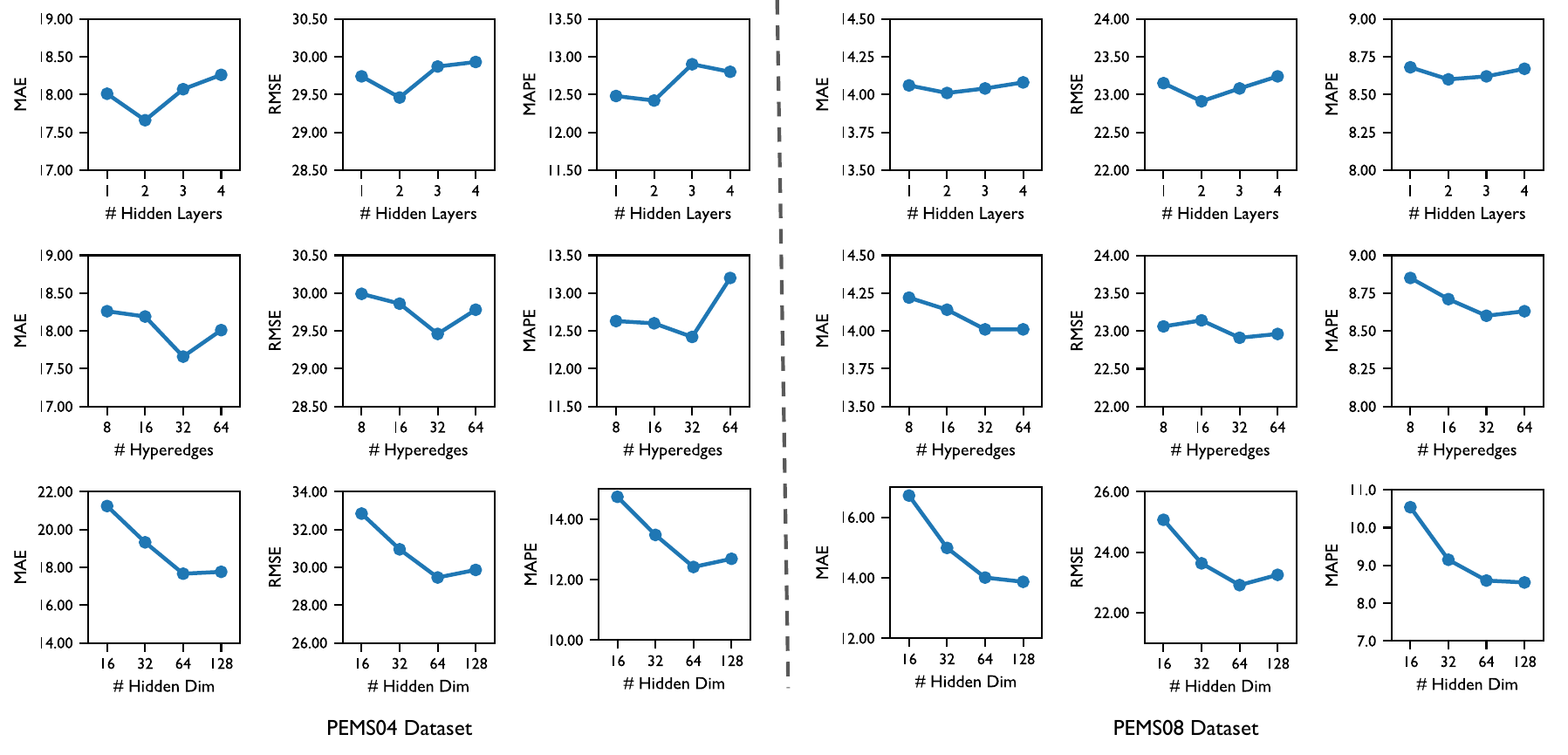}
\caption{The parameter sensitivity experiments of the proposed \method{} with respect to three metrics (MAE, RMSE and MAPE) on two datasets (PEMS04 and PEMS08). The first line studies the number of hidden layers in the Multi-scale Holistic Correlation Extraction module (\emph{i.e.} $L_s$). The second line shows the influence of the number of hyperedges (\emph{i.e.} $I$) in DHSL. The third line focuses on the number of hidden feature dimensions (\emph{i.e.} $d$).
}
\label{fig:sensitivity}
\end{figure*}

\subsection{Ablation Studies}
In this subsection, we conduct ablated studies to evaluate the effectiveness of each proposed Dynamic Hypergraph Structure Learning block, Interactive Graph Convolution block and Multi-scale Holistic Correlation Extraction.

{
\begin{table}[h]
    \centering
    \caption{Ablated study about Dynamic Hypergraph Structure Learning block on PEMS03 and PEMS04 datasets. Different approaches for structure learning (SL) is compared.}
    \label{tab:abl_1}
    \begin{tabular}{c ccc c ccc}
        \toprule
        \midrule  
        \multirow{2}{*}{\textbf{SL}}  &  \multicolumn{3}{c}{\textbf{PEMS03}}    && \multicolumn{3}{c}{\textbf{PEMS04}}\\\cmidrule{2-4} \cmidrule{6-8}
& MAE & RMSE & MAPE && MAE & RMSE & MAPE
\\ \midrule
DHSL & {15.49} & {27.06}    &   {14.38}\% && {17.66}    & {29.46}    &  {12.42}\%
\\ 
NSL & {16.43} & {29.09}    &   {15.21}\% && {18.19}    & {29.88}    &  {13.45}\%
\\ 
FS & {18.91} & {33.85}    &   {20.54}\% && {24.32}    & {40.35}    &  {15.57}\%
\\ \midrule\bottomrule
    \end{tabular}
  \end{table}
}
\noindent\textbf{Dynamic Hypergraph Structure Learning block.} In table~\ref{tab:abl_1}, we perform ablation studies on the Dynamic Hypergraph Structure Learning (DHSL) block of the model by comparing different approaches with respect to structure learning (SL). Line 1 (DHSL) is the original model using DHSL. Line 2 is the model without structure learning (NSL). Comparing Line 1 and Line 2, we can see that the proposed dynamic hypergraph structure learning block is important for traffic prediction, since removing the module lead to a significant performance drop in all metrics. This also suggests that the predefined road network is not perfect. Incomplete or corrupted data might exist, and more importantly, the semantic information beyond the topological proximity is lost.
To solve this problem, a simple approach is to learn the adjacency matrix from scratch (\emph{i.e.} using a learnable adjacency matrix as parameters), which is shown in Line 3 (FS). As we can see, learning the proximity of all the nodes from scratch is catastrophic, and we blame this on the coupling of structure learning and data learning that results in too many parameters without proper supervision. In comparison, the proposed ``low-rank" solution and hypergraph structure learning lead to fewer parameters that are under better supervision.

{
\begin{table}[h]
    \centering
    \caption{Ablated study about the Interactive Graph Convolution (IGC) block on PEMS03 and PEMS04 datasets.}
    \label{tab:abl_2}
    \begin{tabular}{c ccc c ccc}
        \toprule
        \midrule  
        \multirow{2}{*}{\textbf{IGC}}  &  \multicolumn{3}{c}{\textbf{PEMS03}}    && \multicolumn{3}{c}{\textbf{PEMS04}}\\\cmidrule{2-4} \cmidrule{6-8}
& MAE & RMSE & MAPE && MAE & RMSE & MAPE
\\ \midrule
w/ & {15.49} & {27.06}    &   {14.38}\% && {17.66}    & {29.46}    &  {12.42}\%
\\ 
w/o & {16.95} & {29.46}    &   {17.15}\% && {17.99}    & {30.37}    &  {14.13}\%
\\ \midrule
        \bottomrule
    \end{tabular}
  \end{table}
}
\noindent\textbf{Interactive Graph Convolution block.} Table~\ref{tab:abl_2} lists the results of ablation studies with respect to the Interactive Graph Convolution (IGC) block in \method{}. As can be seen from the results, removing the IGC block causes the prediction errors to rise on all metrics, which demonstrates the effectiveness of the proposed mechanism. It is worth noting that without the IGC block, there is a significant increase in RMSE, which weighs more on large errors, and MAPE, which weighs more on errors with small ground truth value (an extreme example would be: if the ground truth traffic flow is 4 and the predicted value is 20, the MAPE of the prediction will be 500\%; if the ground truth is 100 and the predicted value is 116, with the same MAE, the MAPE of this prediction drops to 16\%). On the one hand, the change in RMSE shows that the IGC block can make the prediction more reasonable, avoiding large mistakes to some extent. This suggests that using the high-order interactive features among neighbors yields more robust representations for the center node. On the other hand, the change in MAPE shows that the IGC block can help the model when the traffic flow is lower, which often occurs during sudden external events (\emph{e.g.} car accidents). One possible explanation for this is that learning complex high-order interactions among adjacent locations is helpful for detecting sudden external events and reasoning their influence on the traffic flow.

{
\begin{table}[h]
    \centering
    \caption{Ablated study about multi-scale feature extraction on PEMS03 and PEMS04 datasets.}
    \label{tab:abl_3}
    \begin{tabular}{c ccc c ccc}
        \toprule
        \midrule  
        \multirow{2}{*}{\textbf{\#Scale}}  &  \multicolumn{3}{c}{\textbf{PEMS03}}    && \multicolumn{3}{c}{\textbf{PEMS04}}\\\cmidrule{2-4} \cmidrule{6-8}
& MAE & RMSE & MAPE && MAE & RMSE & MAPE
\\ \midrule
1 & {15.61} & {27.26}    &   {15.28}\% && {18.14}    & {29.95}    &  {12.99}\%
\\ 
2 & {15.54} & {27.17}    &   {14.81}\% && {18.07}    & {29.76}    &  {12.47}\%
\\ 
6 & {15.49} & {27.06}    &   {14.38}\% && {17.66}    & {29.46}    &  {12.42}\%
\\ \midrule
        \bottomrule
    \end{tabular}
  \end{table}
}
\noindent\textbf{Multi-scale Holistic Correlation Extraction.} Table~\ref{tab:abl_3} shows the results in regard to the Multi-scale Holistic Correlation Extraction (MHCE). The first line shows the model's performance using only one scale; the second line shows the performance with two scales (\emph{i.e.} $J=2$ and $\epsilon \in \{1, 3\}$); the third line shows the performance with six scales, which is the design choice of \method{}. As can be seen from the data, increasing the number of scales leads to an improvement in prediction. This suggests that there are traffic patterns with different periodicity, which need to be captured in different granularity. The proposed \method{} extracts features from a variety of scales, which equip the model with the ability to capture patterns with different granularity.

\begin{figure*}[ht]
\centering
\includegraphics[width=0.86\textwidth,keepaspectratio=true]{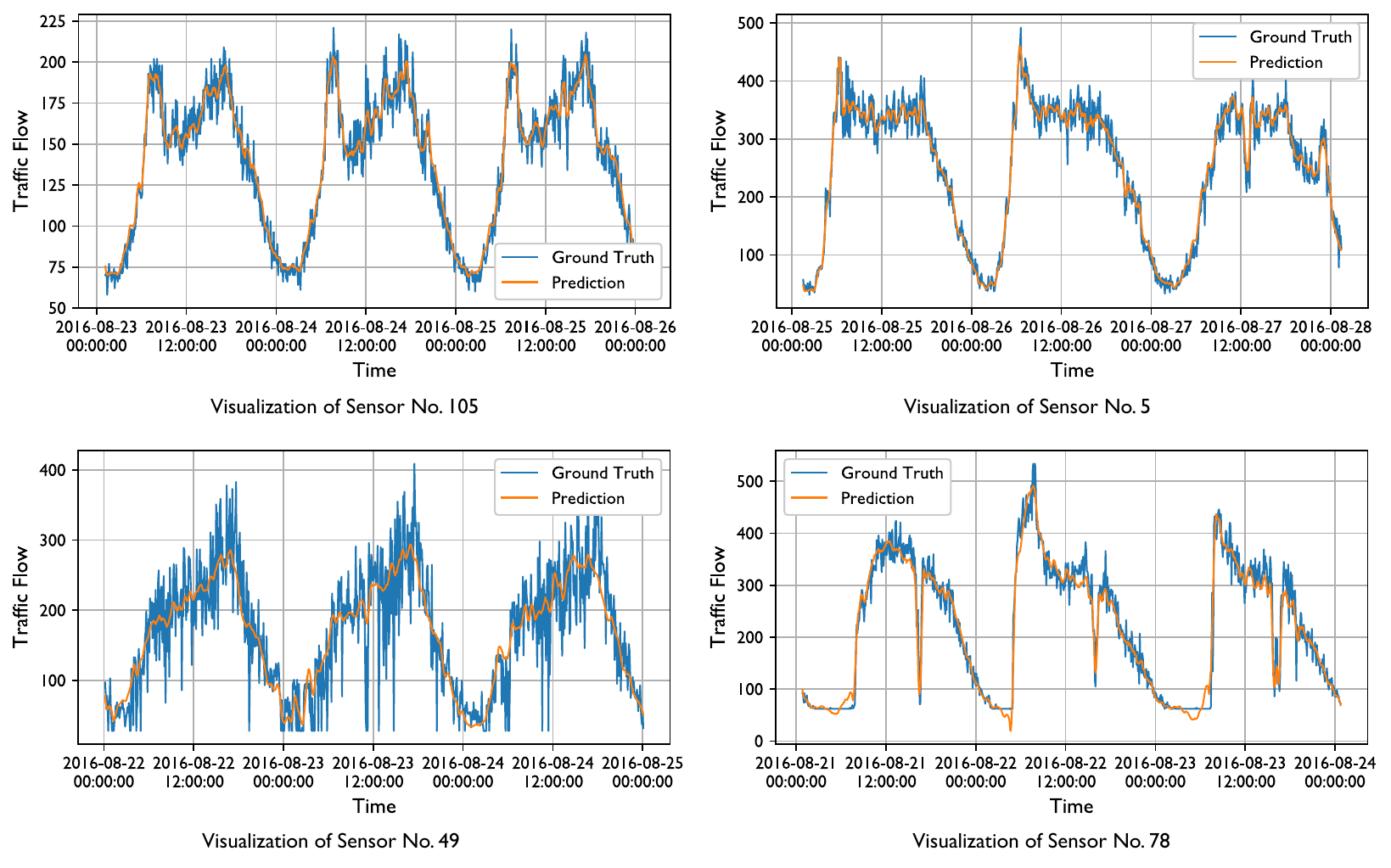}
\caption{Visualization of prediction results on PEMS08 dataset. The results show that the model has the ability to not only capture regular patterns (upper left) but also adapt to pattern changes (upper right). Moreover, the model yield reasonable predictions in adverse cases when there is too much noise (lower left) and strange patterns (lower right).
}
\label{fig:vis}
\end{figure*}

\subsection{Hyperparameter Analysis }\label{sec:hyper}
Here we study the hyperparameters of \method{}. Specifically, we focus on the number of hidden layers (\emph{i.e.} $L_s$) in MHCE, the number of hyperedges (\emph{i.e.} $I$) in the DHSL block and the hidden dimension (\emph{i.e.} $d$) of the feature. The experimental results on two datasets (PEMS04 and PEMS08) are shown in Fig.~\ref{fig:sensitivity}. Note that when studying the effect of one hyperparameter, others are kept as the default values. Our observations and analysis are summarized as follows:
\begin{itemize}
    \item Generally, the proposed \method{} is not sensitive to changes in hyperparameters. As can be seen from the data, changing the number of hidden layers and the number of hyperedges does not influence the performance very much, and in most cases (except the cases when we use small hidden dimensions like 16 and 32), the change in errors (MAE, RMSE, MAPE) is minimal. This shows the robustness of our model.
    We also observe that the model is less sensitive to hyperparameter changes in the PEMS08 dataset, compared to the PEMS04 dataset. A possible explanation for this is that the traffic flow in PEMS08 is easier to predict (the prediction errors in PEMS08 are lower than those in PEMS04).
    \item We vary the number of hidden layers in the range of $\{1, 2, 3, 4\}$. The first line in Fig.~\ref{fig:sensitivity} shows the results. Although the performances do not change too much, the best performance is achieved by 2 hidden layers. One explanation for this result is that the model requires enough layers to enlarge the receptive field and to learn higher-order relations in the structured data. On the other hand, deeper layers introduce extra parameters, which might be harder to learn. Therefore, a moderate number of hidden layers (\emph{i.e.} 2) is adopted in \method{}.
    \item We also perform an experiment with the number of hyperedges ($I$) in the DHSL block. Specifically, we vary the number of hyperedges $I$ in the range of $\{8, 16, 32, 64\}$, and the result is shown in the second line in Fig.~\ref{fig:sensitivity}. As can be seen from the results, using 32 hyperedges is relatively better on the two datasets. A possible reason for this phenomenon is that a small number of hyperedges only allow the model to capture coarse information in the graph, which hinders prediction accuracy. On the contrary, too many hyperedges bring extra structural correlations, which might introduce unnecessary noise to the model. Thus, an intermediate number of hyperedges is more helpful for the model.
    \item For the number of hidden dimensions ($d$), we vary the value in the range of $\{16, 32, 64, 128\}$, and the result is shown in the third line in Fig.~\ref{fig:sensitivity}. From the data, we can see that the model performs badly when the number of hidden dimensions is very small, which affects the model's ability to capture complex spatio-temporal dynamics. On the other end, we observe that there is no significant performance gain when the number of hidden dimensions grows beyond 64. Therefore, we set the hidden dimension of our model as 64.
\end{itemize}

\begin{figure*}[ht]
\centering
\includegraphics[width=0.9\textwidth,keepaspectratio=true]{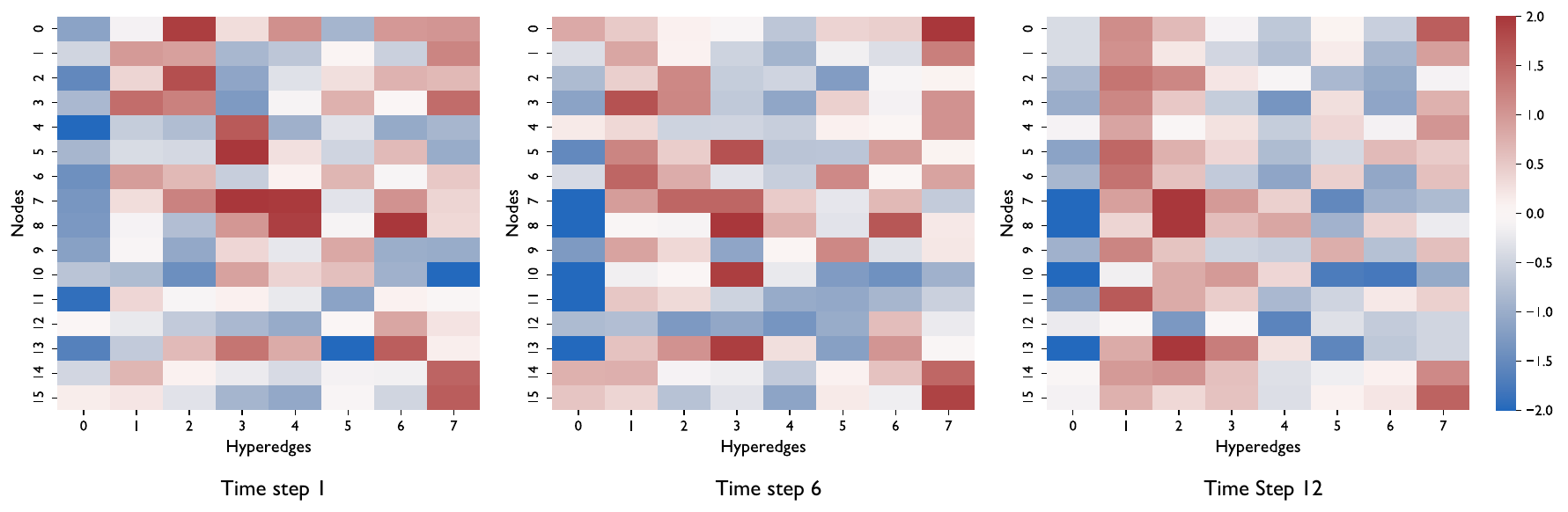}
\caption{Visualization of hypergraph incidence matrix (\emph{i.e.} $\bm{\Lambda}$). The experiment is performed on the PEMS08 dataset, and we extract three time steps in the feature (time step 1, 6 and 12). For better visualization, we only show the submatrices of the incidence matrix.
}
\vspace{-1mm}
\label{fig:heatmap}
\end{figure*}

\subsection{Case Study}
In this subsection, we visualize four prediction results in comparison to the ground truth data on the PEMS08 dataset, which is shown in Fig.~\ref{fig:vis}. The upper-left figure is the prediction results of Sensor 105, from August 23 to 25 in 2016. The three days are all workdays. As can be seen from the results, the daily patterns are similar, and the model can learn these easily. The upper-right figure is the prediction results of Sensor 5, from August 25 to 27 in 2016. The first two days are workdays while the last day is Saturday. As we can see, the last day's traffic pattern is different from the first two days. Despite this pattern change, the model can adapt to this nicely. For example, around 12 a.m. on August 27, the traffic flow suddenly decreases, and our model quickly adapt to the change and predict lower traffic flow.

The lower-left figure is the prediction results of Sensor 49, from August 22 to 25. Although the three days are all workdays, there is much noise in the traffic signal, which makes the prediction task challenging. As we can see from the prediction results, our model yields reasonable predictions, which shows that the model is robust to noises in the traffic flow signal. The lower-right figure is the prediction results of Sensor 78, from August 21 to 23. This sensor exhibits a strange pattern: at night the traffic flow drops to somewhere around 60 and then becomes stable around that point. This could be a defect in the data, or it might result from some regular events like police patrol. Although our model does not yield a constant value for these time periods, this could be a shortcoming among most data-driven methods.

\subsection{Analysis of the Dynamic Hypergraph Structure Learning}
In this subsection, we provide some further analysis of the Dynamic Hypergraph Structure Learning block in our model. Fig.~\ref{fig:heatmap} shows the visualization results of the incidence matrix (\ie $\bm{\Lambda}$ in Eq.~\ref{eq:incidence}). Remember that the entries in the incidence matrix denote the closeness between a node and a hyperedge: the larger the value, the closer a node is to the hyperedge. For better visualization, the figure only shows submatrices of the incidence matrix. From the results, we can see that different nodes have different degrees of closeness to different hyperedges. For example, in the left matrix, Node 0 and 2 are closer to Hyperedge 2, whereas Node 4 and 5 are closer to Hyperedge 3. This shows that different nodes are linked by different hyperedges so that we can learn the relations among multiple nodes on the hypergraph more efficiently.

Moreover, the closeness of a node and a hyperedge changes over time, which shows that hypergraph learning can capture the spatio-temporal dynamics in the traffic data. Fig.~\ref{fig:heatmap} shows three time steps (time step 1, time step 6 and time step 12) in the temporal graph data. As we can see, the closeness between nodes and hyperedges evolves over time. For example, at time step 1, Node 0 is closely related to Hyperedge 2. However, in time step 6, the node is leaving Hyperedge 2 and joining Hyperedge 7. \R{This shows that our model can capture a shift of influence in the traffic network across time. For instance (also illustrated in Fig.~\ref{fig:motivation}), a node near the residential area may exhibit similar patterns with other nodes inside the residential area (and thus they are closer to Hyperedge A, influenced more by this hyperedge). However, a car accident nearby could change the future traffic pattern of this node, and the node tends to leave Hyperedge A and join Hyperedge B, which represents the impact of the car accident.}

Interestingly, some hyperedges exhibit functions similar to other components commonly used in traffic flow prediction (or more generally, multi-variate time series forecasting). For example, Hyperedge 1 at time step 12 connects to most nodes, which functions like a spatial aggregation at some time step. For another example, Hyperedge 6 at time step 12 is close to some nodes and distant from others. When the node features aggregate to this hyperedge, some features are multiplied by positive values and others are multiplied by negative values, which indicates that this hyperedge is performing convolution operation on the node features, and that its functionality is similar to graph convolutions.

\section{Conclusion}
\label{conclusion}

This paper proposes a novel model named \method{} for traffic flow forecasting, which models both non-pairwise and high-order relationships in the traffic network. 
To describe non-pairwise dynamic interactions, we offer a block for constructing a temporal hypergraph in which all nodes are observations at every timestamps. Then, we develop hypergraph convolution to update node representations using data from their related hyperedges. In addition, to investigate high-order spatio-temporal interactions in the road network, an interactive graph convolution block is introduced. Finally, we combine these two block into a comprehensive multi-scale correlation extraction framework. One limitation of our model is that although we have tried to reduce the parameters, it might also have the risk of overfitting. 
\section*{Acknowledgments}
This paper is partially supported by grants from the National Key Research and Development Program of China with Grant No. 2018AAA0101902 and the National Natural Science Foundation of China (NSFC Grant Number 62276002).

\balance
\bibliographystyle{IEEEtran}
\bibliography{mybibfile}

\end{document}